\documentclass[sigconf,screen]{acmart}
\usepackage{multicol}
\usepackage{multirow}
\usepackage{array}
\usepackage{graphicx}
\usepackage{subcaption}
\usepackage{caption}
\usepackage{enumitem}

\AtBeginDocument{%
  }

\renewcommand\footnotetextcopyrightpermission[1]{}
\begin{document}

\title{Learning Adaptive Node Selection with External Attention for Human Interaction Recognition}


\author{Chen Pang}
\affiliation{%
  \institution{Shandong Normal University}
  \country{China}}
\email{chenp0721@163.com}

\author{Xuequan Lu}
\affiliation{%
  \institution{University of Western}
  \country{Australia}}
\email{bruce.lu@uwa.edu.au}

\author{Qianyu Zhou}
\affiliation{%
  \institution{Jilin University}
  \country{China}}
\email{zhouqianyu@sjtu.edu.cn}

\author{Lei Lyu}
\affiliation{%
  \institution{Shandong Normal University}
  \country{China}}
\email{lvlei@sdnu.edu.cn}








\begin{abstract}
Most GCN-based methods model interacting individuals as independent graphs, neglecting their inherent inter-dependencies. Although recent approaches utilize predefined interaction adjacency matrices to integrate participants, these matrices fail to adaptively capture the dynamic and context-specific joint interactions across different actions. 
In this paper, we propose the Active Node Selection with External Attention Network (ASEA), an innovative approach that dynamically captures interaction relationships without predefined assumptions. 
Our method models each participant individually using a GCN to capture intra-personal relationships, facilitating a detailed representation of their actions. To identify the most relevant nodes for interaction modeling, we introduce the Adaptive Temporal Node Amplitude Calculation (AT-NAC) module, which estimates global node activity by combining spatial motion magnitude with adaptive temporal weighting, thereby highlighting salient motion patterns while reducing irrelevant or redundant information. A learnable threshold, regularized to prevent extreme variations, is defined to selectively identify the most informative nodes for interaction modeling.
To capture interactions, we design the External Attention (EA) module to operate on active nodes, effectively modeling the interaction dynamics and semantic relationships between individuals. 
Extensive evaluations show that our method captures interaction relationships more effectively and flexibly, achieving state-of-the-art performance. \textit{Our code will be made publicly available}.
\end{abstract}



\keywords{Human Interaction Recognition, Graph Convolution Network, Attention Mechanism}

\maketitle

\section{Introduction}
Human Interaction Recognition (HIR) has emerged as a critical area of research in multimedia analysis, with a primary focus on identifying and understanding human actions within video sequences \cite{51,52}. With the rapid advancements in technologies such as social media, intelligent surveillance, and virtual reality, there is an increasing demand for real-time analysis of human action. Unlike individual action recognition, which focuses on a single subject, HIR centers on interactions between multiple participants. It encompasses two key aspects: the actions of a single subject and the interaction between participants. Understanding the connection between these two aspects is essential for gaining a comprehensive insight into human action. Compared to RGB videos, skeleton-based representation, which models the human body through joint coordinates, is computationally more efficient and less susceptible to occlusions and background variations. Additionally, advancements in skeleton capture devices \cite{19} and pose estimation techniques \cite{18} have made it more cost-effective to obtain precise joint coordinates. As a result, action recognition based on skeleton data has gained significant attention and research in recent years.

\begin{figure*}[]
	\centering
	\includegraphics[width=0.9\textwidth]{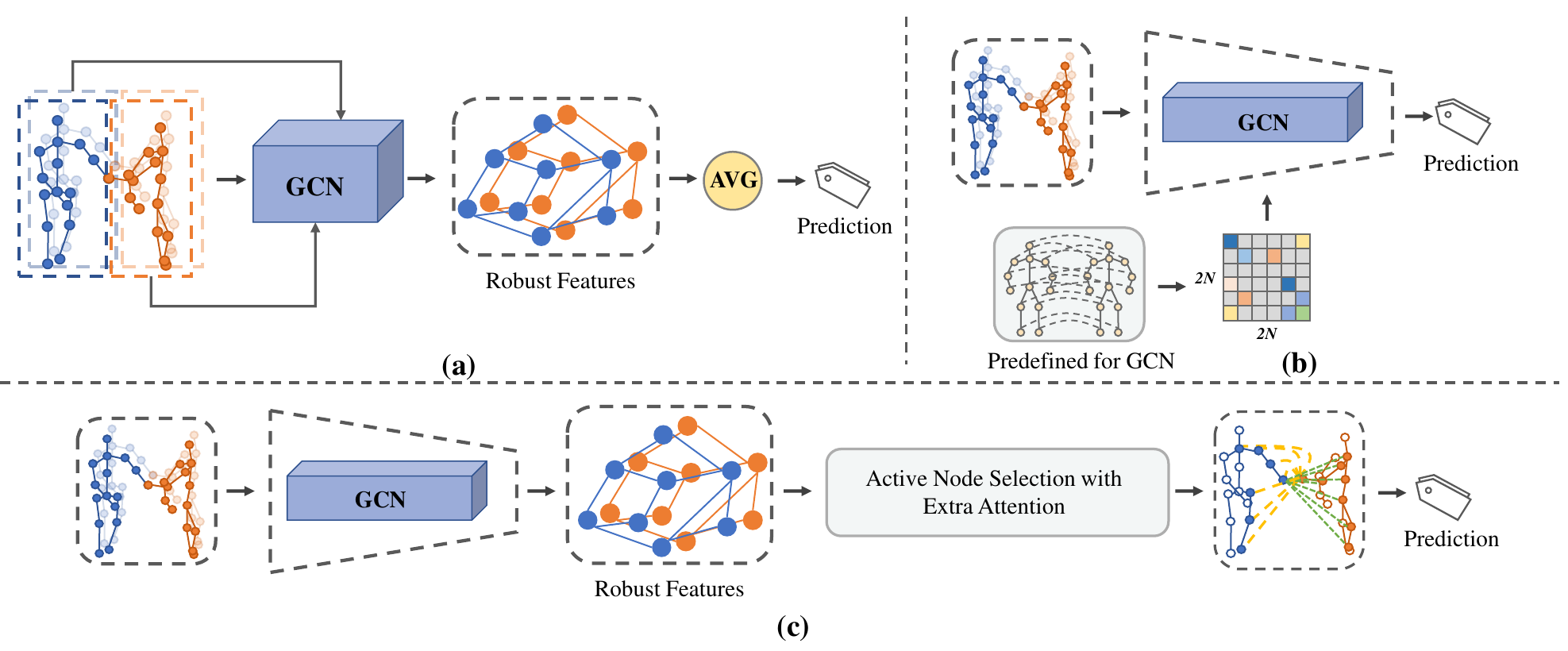}
    \vspace{-3mm}
	\caption{Existing methods struggle with multi-person interactions and inefficient modeling. (a) Most GCN-based methods treat multi-person skeletons as isolated individuals, neglecting interaction dynamics. (b) Recent approaches introduce two-person graph structures with predefined interaction adjacency matrices, which constrain the flexibility of interaction modeling. (c) Our framework dynamically captures interaction relationships by selectively attending to active nodes without relying on predefined assumptions.}
	\label{figvs}
    \vspace{-3mm}
\end{figure*}

GCN-based methods \cite{16,17,20} have demonstrated strong performance in skeleton-based action recognition by effectively capturing spatial dependencies in non-Euclidean data. However, these methods  \cite{45,46} often treat multi-person skeletons as independent individuals, neglecting crucial interaction dynamics (Fig. \ref{figvs} (a)). To address this limitation, recent approaches \cite{1,3} introduce two-person graph structures, where predefined adjacency matrices are designed to explicitly model interactions (Fig. \ref{figvs} (b)). While these methods enhance interaction representation, their fixed structures impose rigid constraints, limiting adaptability to diverse interaction patterns and potentially introducing redundant connections that degrade representation quality.
To improve flexibility, some methods incorporate knowledge-informed or adaptive learning-based graphs. For instance, K-GCN \cite{2} employs both knowledge-given and knowledge-learned graphs to refine joint connectivity, while DR-GCN \cite{4} constructs interaction adjacency matrices based on joint distances and similarity measures. However, these approaches primarily rely on geometric relationships, which may be insufficient for interactions conveyed through gaze, gestures, and facial expressions, which are common in scenarios such as photography.
 
To address these challenges, attention mechanisms have been introduced to enhance interaction modeling. IGFormer \cite{5} transforms skeleton sequences into body-part sequences, enabling regions to attend to others at both semantic and spatial levels. And ISTA-Net \cite{6} unfolds entities along the channel dimension and applies attention to partitioned features via a sliding window. Although these methods improve interaction modeling, global attention mechanisms may introduce redundancy, obscure critical interaction patterns, and increase computational complexity, leading to reduced robustness in noisy environments. These limitations highlight the need for a more adaptive and selective strategy for multi-person interaction modeling.

In this paper, we propose the Active Node Selection with External Attention Network (ASEA), which dynamically captures interaction relationships without predefined assumptions. 
Our key idea is to enhance interaction modeling by selectively attending to the most contextually relevant nodes that are closely associated with the interaction actions, thereby suppressing redundant or irrelevant information (Fig. \ref{figvs} (c)).
This adaptive strategy facilitates more flexible and context-aware modeling, improving both representational expressiveness and computational efficiency.
Specifically, we first utilize a GCN to model each participant individually, capturing intra-personal relationships for a detailed representation of actions. These individual action representations lay a solid foundation for capturing meaningful inter-personal interactions. To identify active nodes involved in interactions, we propose the Adaptive Temporal Node Amplitude Calculation (AT-NAC) module, which computes global node activity by combining temporal attention weights derived from node feature statistics with the $L_2$-norm of node features. This enables the model to emphasize informative segments while suppressing noise across time. In addition, a learnable threshold is defined to adaptively select active nodes, and regularization is introduced to stabilize this thresholding process. To capture semantic correlations among individuals while reducing redundancy, we propose an External Attention (EA) module, which is applied to the selected active nodes. This module enhances the modeling of inter-personal relationships by attending to the most informative representations. This approach ensures that the most informative motion cues are emphasized, improving both efficiency and interpretability in interaction modeling.

 The main contributions of this paper are as follows:

\vspace{10pt}
\begin{itemize}[topsep=2pt,itemsep=2pt,leftmargin=1.5em]

\item 
We propose the Active Node Selection with External Attention Network (ASEA), which dynamically selects informative joints and applies the proposed external attention to capture critical inter-individual relationships. This adaptive strategy eliminates the need for predefined interaction assumptions, enabling more flexible and context-aware interaction modeling.

\item 
We introduce the Adaptive Temporal Node Amplitude Calculation (AT-NAC) module to quantify joint activity by integrating variance-based temporal weighting with $L_2$-norm feature amplitudes. It adaptively selects active nodes by adjusting the selection threshold based on amplitude distribution, emphasizing high-intensity frames while suppressing noise. To stabilize thresholding and ensure consistency across interactions, we introduce a regularization constraint to mitigate parameter drift.
\item 
To capture inter-individual interactions and semantic dependencies while mitigating unnecessary interaction noise, we design an external attention mechanism that dynamically operates on selected active nodes.
\end{itemize}

\section{Related Work}
Skeleton data inherently exists in the form of a topological graph, and graph convolution is a deep learning method based on graph structures that efficiently extracts and classifies features from human skeleton data. Therefore, graph convolution-based human action recognition has become the mainstream approach for processing skeleton data.

\subsection{Action recognition}
Yan et al. \cite{8} first applied graph convolutional networks to skeleton sequence modeling, proposing a Spatio-Temporal Graph Convolutional Network (ST-GCN) that constructs a spatio-temporal graph reflecting the natural connections of human joints. The graph topology is manually defined, and remains fixed across all network layers and for different input samples. To address this limitation, Shi et al. \cite{9} introduced a two-stream adaptive GCN (2s-AGCN). Building upon the adjacency matrix and mask matrix of ST-GCN, they incorporated a unique graph for each sample and replaced the direct multiplication of the adjacency matrix and mask matrix in ST-GCN with matrix addition, effectively constructing non-existent connections. Further enhancing flexibility, Shi et al. \cite{10} extended 2s-AGCN by embedding an attention module into each convolutional layer, allowing the model to more dynamically focus on different joints, frames, and channels. Recognizing that different types of motion features correspond to varying relationships between joints, CTR-GCN \cite{11} uses a shared adjacency matrix as a prior across all channels, allowing the network to learn specific relationships for each channel. This approach improves the model's ability to capture topological information by differentiating the topology both between layers and channels. To capture context-dependent topologies, InfoGCN \cite{12} leverages an information bottleneck objective and attention-based graph convolution, while incorporating a multi-modal skeleton representation based on joint relative positions to enhance spatial information for better classification. Further advancing the model's capacity to capture complex motion patterns, Wang et al. \cite{13} proposed a multi-order multi-modal transformer called 3Mformer, which simulates higher-order hyperedges between body joints by forming a hypergraph, thereby capturing higher-order motion patterns of joint groups. To address the challenge of varied representations resulting from different execution styles of the same action, DeGCN \cite{14} introduces a novel deformable graph convolutional network to adaptively select informative joints for skeleton-based action recognition. It learns deformable sampling locations on both spatial and temporal graphs, allowing the model to capture discriminative receptive fields. With the development of vision-language model, CFVL \cite{15} integrates it into skeleton-based action recognition. It uses a feedback decoder to align language representations with skeleton frames and employs an adaptive projection module to extract spatiotemporal features, ensuring effective alignment of vision and language information. However, the aforementioned methods primarily model individuals independently, overlooking critical interaction relationships, which limits their effectiveness in complex interactive action recognition tasks.

\vspace{-2mm}
\subsection{Interactive Action Recognition}
Recently, several new networks have been proposed to address this issue, aiming to better capture the interactions between individuals. Yang et al. \cite{1} introduced a pairwise graph by incorporating inter-body connections into the isolated two-person graph. They further extended the ST-GCN model to ST-GCN-PAM, leveraging pairwise graph convolution to capture interactions between the two bodies. Li et al. \cite{3} designed a unified two-person graph to capture both inter-body and intra-body correlations, addressing the limitations of existing methods that treat two-person interactions as separate graphs. It employs various graph labeling strategies to enhance spatial-temporal feature learning, improving the recognition of both interactions and individual actions. However, these predefined adjacency matrices impose rigid structural assumptions, limiting their adaptability to diverse interaction patterns. K-GCN \cite{2} introduced two novel graph structures: a knowledge-given graph for direct connections between individuals and a knowledge-learned graph for dynamically capturing the relations between inter-body joints. These two graphs combined with a naturally connected graph are send to the knowledge-embedded graph convolutional network to improve interaction recognition. Zhu et al. \cite{4} proposed a GCN model that separately handles intra-body and inter-body relationships. For inter-body interactions, they introduced a dynamic relational adjacency matrix, combining joint inverse distance and similarity to capture the connection between individuals. The results from both intra-body and inter-body graph convolutions are then summed to obtain the final output, improving two-person interaction recognition. IGFormer \cite{5} is the first network to adopt a Transformer-based architecture for skeleton-based interaction recognition, utilizing prior knowledge of human body structure to design interaction graphs. It constructs these graphs based on semantic and distance correlations between interacting body parts, enhancing each person's representation by aggregating information from the interacting body parts. AIGCN \cite{7} encoded skeleton data with two semantic vectors to generate a dynamic adjacency matrix using MLPs and an attention mechanism. They also defined a static adjacency matrix to represent simple interactive relationships, based on the mirror symmetry of joints. ISTA-Net \cite{6} eliminates the need for subject-specific graph prior knowledge, capturing interactive and spatiotemporal correlations by extending the entity dimension along the channel. It then uses attention mechanisms with a sliding window to effectively model spatial, temporal, and interactive relationships. Although \cite{4} performed the intra-body and inter-body graph convolutions separately, the results from both are simply summed to obtain the final output. This operation may lead to the loss of important interaction information, especially in complex interaction scenarios where the relationships can be subtle and intricate, and it fails to account for the dynamic changes in individual features and interaction relationships. In our work, we first model individual actions to capture dynamic characteristics. Based on this, we attempt to select the most relevant active nodes that reflect key interactions. Then, we leverage external attention to these active nodes to enhance interaction modeling. The modeling of individual actions guides the selection of active nodes, which then serves as the foundation for applying attention, ensuring precise and effective interaction capture.  

\begin{figure*}[h]
	\centering
	\includegraphics[width=0.92\textwidth]{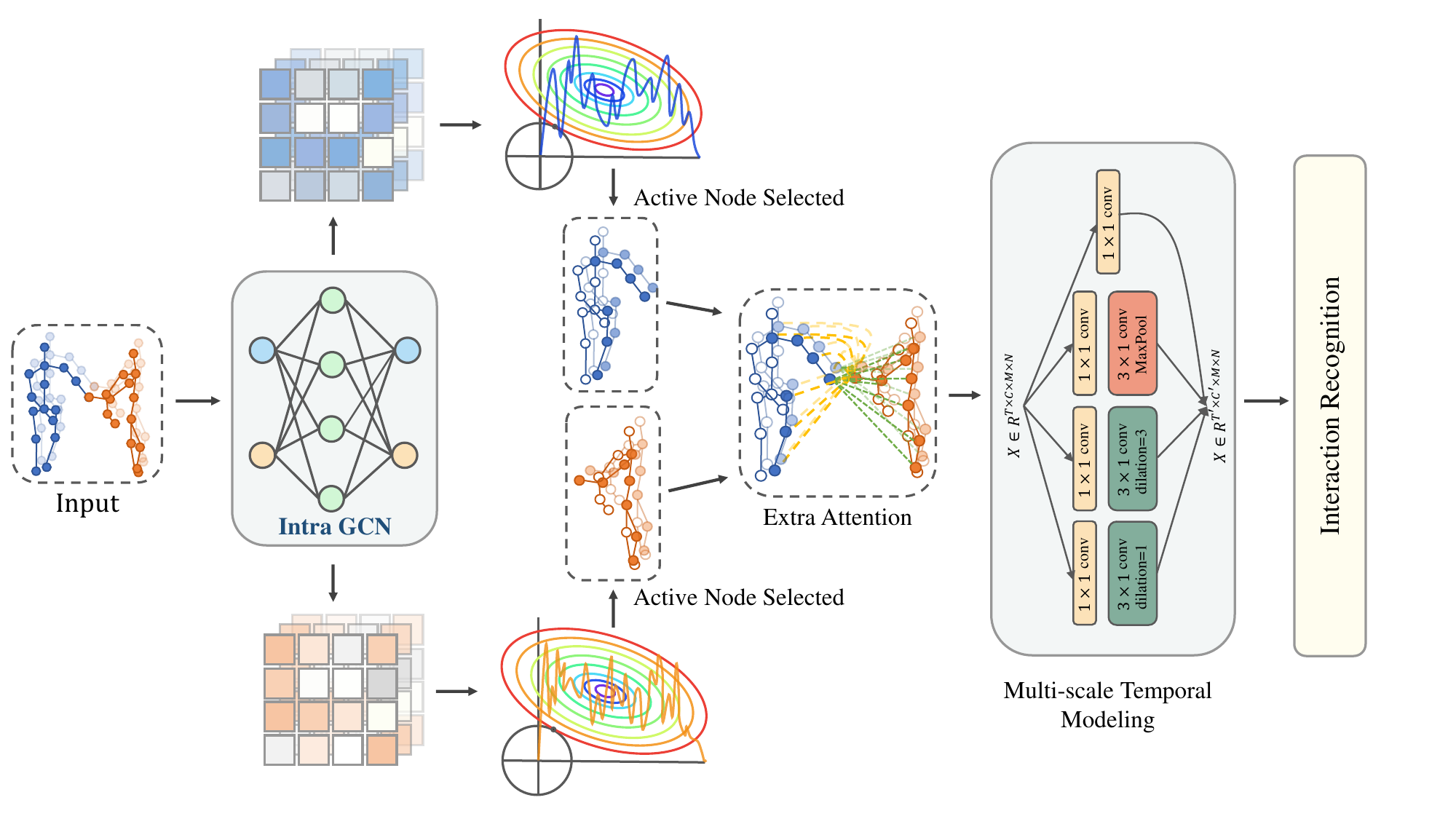}
    \vspace{-5mm}
	\caption{The overview of our Active Node Selection with External Attention Network (ASEA). Given an input skeleton sequence, the Intra GCN encoder first captures individual action features. Subsequently, the Adaptive Temporal Node Amplitude Calculation (AT-NAC) module identifies active nodes by jointly modeling their spatial characteristics and temporal dynamics. These selected active nodes are then processed by the External Attention (EA) module, which applies an attention mechanism across individuals to effectively capture inter-personal interactions. Finally, the aggregated representation is fed into a multi-scale temporal modeling module to learn temporal dependencies at different scales.}
	\label{fig.1}
    \vspace{-3mm}
\end{figure*}

\section{Method}
\subsection{Problem Definition}

The skeleton sequence is denoted as $X \in R^{C \times T \times M \times N}$, where $C$ represents the coordinate dimension, $T$ represents the total number of frames, $N$ represents the number of joints, and $M$ is the number of people. Most previous studies consider the two people as a whole and organize the input skeleton data as $X \in R^{C \times T \times MN}$ to model the intra- and inter-body connections with a single adjacency matrix $A \in R^{MN \times MN}$, where interaction relationships are typically predefined. In contrast, our approach avoids these manually defined interaction models by adopting a more flexible and data-driven strategy. Specifically, we first model the action of each individual based on the adjacency matrix $A \in R^{N \times N}$, capturing the individual action features $X \in R^{C \times T \times N}$. Then, we identify the active nodes $S\in R^N$ (i.e., the individuals involved in the interaction) based on their action features, considering both spatial and temporal dimensions. After selecting the active nodes, we apply an external attention mechanism to model the interactions between these nodes. The attention mechanism allows us to capture the dependencies between active individuals by dynamically focusing on their relevant interactions.

\subsection{Overview}
The overall architecture of the proposed Active Node Selection with External Attention Network (ASEA) is illustrated in Fig. \ref{fig.1}. Given a skeleton sequence containing two interactive subjects, it is fed into the GCN encoder to capture individual action features. Building on these extracted features, the Adaptive Temporal Node Amplitude Calculation (AT-NAC) module models both the spatial characteristics of the nodes and their temporal variations to accurately identify active nodes.
The AT-NAC module generates temporal attention weights based on the statistical properties of node features in each frame, thereby enhancing the contribution of key action segments while minimizing the impact of low-information or noisy frames. These adaptive temporal weights are then combined with the $L_2$-norm of the node features to compute the global node activity, enabling the selection of the most relevant nodes and capturing their spatiotemporal dynamic features. Then, the External Attention (EA) module applies an attention mechanism on the active nodes across different individuals, capturing their semantic correlations and effectively modeling the inter-personal interactions. Finally, the aggregated representation is fed into a multi-scale temporal modeling module to learn temporal dependencies at different scales.

\subsection{Intra Graph Convolution Network}
\textbf{Spatial Modeling.} We employ ST-GCN with channel-wise topology \cite{11} as a baseline, optimizing it by considering the specific correlations within each channel to generate a channel-level topology that more accurately reflects the underlying motion characteristics. In channel-wise topology, the parameterized adjacency matrix $A \in R^{N \times N}$ is shared across all channels and optimized through backpropagation. Meanwhile, the channel-specific correlation $Q\in R^{N\times N\times C^\prime}$ is learned to model specific relationships between vertices in $C^\prime$ channels. Specifically, for a pair of vertices ($v_i$, $v_j$) with their corresponding features ($x_i$, $x_j$), linear transformations $\partial$ and $\varphi$ are used to reduce the feature dimension. Then, the distance between $\partial(x_i)$ and $\varphi(x_j)$ along channel dimension is calculated and the nonlinear transformations of these distance is utilized as channel-specific topological relationship between $v_i$ and $v_j$.  
\begin{equation}
    q_{i j}=\delta\left(\sigma\left((x_i)-\varphi(x_j)\right)\right),
\end{equation}
where $\sigma(\cdot)$ is activation function and $\delta$ denotes linear transformation. $q_{i j} \in R^{C^{\prime}}$ is a vector in $Q$.

The channel-wise topologies $R \in \mathbb{R}^{N \times N \times C'}$ are derived by enhancing the uniform topology $A$ with channel-specific correlations $Q$:
\begin{equation}
    R=A+\alpha Q,
\end{equation}
where $\alpha$ represents a trainable scalar that modulates the level of refinement. The addition is executed through broadcasting, with $A$ being added to each channel of $\alpha Q$.


\noindent \textbf{Temporal modeling.} To model actions with varying durations, we design a multi-scale temporal modeling module inspired by the approach in \cite{20}. Unlike the original design, we utilize fewer branches to avoid excessive computational overhead, as too many branches can significantly slow down inference speed. As shown in Fig. \ref{fig.1}, this module contains four branches, each containing a $1\times1$ convolution to reduce channel dimension. Each of the first three branches contains two temporal convolutions with different dilation rates, followed by a Max-Pooling operation. The outputs from all four branches are then concatenated to produce the final result. 

\subsection{Adaptive Temporal Node Amplitude Calculation}
In interactive action recognition, the key interactions are typically driven by a small subset of nodes between the participants. For instance, in shaking hands, the hands and arms of the individuals are the primary nodes responsible for conveying the core information about the interaction. By identifying and focusing on these active nodes, the model can more effectively capture the dynamic and intentional aspects of the interaction between individuals. This approach can help avoid the unnecessary modeling of non-essential information, thereby improving the model's performance in complex scenarios by emphasizing the most informative features. In this work, we propose the Adaptive Temporal Node Amplitude Calculation (AT-NAC) module to account for the spatial characteristics of the nodes (such as the movement amplitude of key joints) and the variations along the temporal dimension.

The skeleton sequence features obtained from the intra-GCN are represented as $X\in R^{B\times C\times T\times N}$, where $B$ denotes the batch size, $C$ represents the feature channels, $T$ is the number of temporal frames, and $N$ is the number of body joints. For each temporal frame $t$, we compute the joint-wise feature energy using the $L_2$-norm, defined as:
\begin{equation}
  E_{b, t, n}=\left\|X_{b,:, t, n}\right\|_2,
\end{equation}
where $E_{b,t,n}$ quantifies the activity intensity of joint $n$ at frame $t$. This energy measure captures the motion characteristics of individual joints over time, reflecting both their intensity and variability. However, assigning equal weight to all frames can lead to the dilution of important temporal information, especially when certain frames contain significant motion, while others are relatively static. To mitigate this issue and prioritize frames with more dynamic and informative motion, we calculate the variance of joint energy across all joints at each time step, as follows:
\begin{equation}
V_{b, t}=\frac{1}{N} \sum_{n=1}^N\left(E_{b, t, n}-\mu_{b, t}\right)^2, \mu_{b, t}=\frac{1}{N} \sum_{n=1}^N E_{b, t, n},
\end{equation}
where $V_{b,t}\in R^{B\times T}$ measures the dispersion of joint activities in frame $t$. Higher variance indicates significant local motion variations (e.g., sudden hand movements), while lower variance corresponds to stable or globally consistent motions.

To emphasize frames with higher temporal significance, we assign attention weights to the temporal frames by applying the softmax function to the variance values:
\begin{equation}
W_{b, t}=\frac{\exp \left(\gamma \cdot V_{b, t}\right)}{\sum_{t^{\prime}=1}^T \exp \left(\gamma \cdot V_{b, t^{\prime}}\right)}, \gamma>0,
\end{equation}
where $\gamma$ is a temperature parameter to control the sharpness of weight distribution. 
The final joint amplitude for each body joint is computed by aggregating the energy values across the temporally weighted frames:
\begin{equation}
S_{b, n}=\sum_{t=1}^T W_{b, t} \cdot E_{b, t, n}.
\end{equation}
The amplitude $S_{b,n}\in R^{B\times N}$ represents the global activity level of joint $n$, incorporating both spatial intensity and temporal significance. This fusion step effectively captures the joint's contribution to the overall motion dynamics over time. 

To further enhance the model's focus on the most informative regions, it is crucial to identify active joints that dominate the overall movement. However, manually defining a fixed threshold or node count for selecting active joints often leads to suboptimal performance, as different interaction types exhibit varying sparsity levels (e.g., dense full-body motions versus sparse hand gestures). To overcome this limitation, we introduce a learnable statistical criterion for adaptive node selection. For each sample, the mean $\mu_b=\frac{1}{N} \sum_{n=1}^N S_{b, n}$ and standard deviation $\sigma_b=\sqrt{\frac{1}{N} \sum_{n=1}^N\left(S_{b, n}-\mu_b\right)^2}$ of the node amplitudes are calculated.

A high mean indicates overall activity, and a high standard deviation means significant differences between nodes. And we compute sample-specific thresholds based on the distribution of $S_{b,n}$:
\begin{equation}
  \tau_b=\mu_b+\alpha \cdot \sigma_b,
\end{equation}
where $\alpha$ is a trainable parameter initialized to 0.5, which controls the sensitivity of the threshold to the dispersion of the distribution. Lower $\alpha$ values retain subtle local motions (e.g., finger gestures), while higher values focus on dominant joints in coordinated actions (e.g., hugging). Nodes with amplitudes exceeding $\tau_b$ are retained as active joints, and a redundancy safeguard ensures at least one node is selected by preserving the highest-amplitude joint when no candidates meet the threshold.

Finally, we select the nodes whose amplitudes exceed the threshold:
\begin{equation}
    M_{b, n}= \begin{cases}1 & \text { if } S_{b, n}>\tau_b \\
    0 & \text { otherwise }\end{cases},
\end{equation}
These selected active joints are then forwarded to the extra attention module for interaction modeling, where they will serve as the primary nodes for capturing dynamic inter-body interactions.

\subsection{Extra Attention}
Building upon active node selection, we introduce the External Attention (EA) module to effectively capture semantic correlations among these nodes across individuals.
For each active node $n_i^t$ of every individual skeleton graph in the frame $t$, a query vector $q_i^t$, a key vector $k_i^t$, and a value vector $v_i^t$ can be calculated by the trainable linear transformations with three parameters matrices $W_q\in R^{C_{in}\times d_q}$, $W_k\in R^{C_{in}\times d_k}$, $W_v\in R^{C_{in}\times d_v}$ which are shared by all nodes, $q_{p i}^t=W_q x_{p i}^t$, $k_{p i}^t=W_k x_{p i}^t$, and $v_{p i}^t=W_v x_{p i}^t$, where $x_{pi}^t$ represents the feature of the $i$-th node of the $p$-th individual in the $t$-th frame. 

With these vectors, we use the attention to calculate the attention score to weight the value vector $v_{pi}^t$ that corresponds to the query vectors. Taking the example of an interaction involving two individuals, the query vector $q$ corresponding to the first individual interacts with the key vectors $k$ of the second individual, and conversely, the query vectors of the second individual are employed to interact with the key vectors of the first individual, all within the same frame $t$. This structure ensures that the query vector for a given node in one individual interacts with the keys of all active nodes from the other individual to capture their mutual dependencies. The attention scores are computed by performing a dot product between the query vector of the first individual $q_1$ and the key vectors $k_2$ from the second individual, and similarly, the query vectors from the second individual $q_2$ are compared with the key vectors from the first individual. These scores represent the correlation between each pair of nodes across individuals. Specifically, the attention map for each individual is computed as:
\begin{equation}
\alpha_1=\operatorname{softmax}\left(\frac{q_1 \cdot k_2^T}{\sqrt{d_q}}\right), \alpha_2=\operatorname{softmax}\left(\frac{q_2 \cdot k_1^T}{\sqrt{d_q}}\right),
\end{equation}
where $d_q$ is the dimensionality of the query vectors, and the softmax operation ensures that the attention values are normalized, representing the relative importance of the interactions between nodes in different individuals.

Then the attention maps are applied to the corresponding value vectors $v_p$ to aggregate information, 
\begin{equation}
    {out}_1=\alpha_1 \cdot v_2, {out}_2=\alpha_2 \cdot v_1.
\end{equation}

Finally, a residual connection is applied to the output features to integrate the transformed features with the original input, preserving key spatial information while allowing the attention mechanism to refine the representation. 

After capturing the inter-individual interaction relationships, the features of the two individuals are concatenated, resulting in a tensor of shape $(B, C, T, J, M)$, where $B$ denotes the batch size, $C$ represents the number of channels, $T$ is the number of frames, $J$ is the number of active joints, and $M$ is the number of individuals. This concatenated tensor is then fed into the multi-scale temporal modeling module to capture the temporal dynamics across different scales, effectively encoding the temporal dependencies between frames for their interactions over time.

\subsection{Loss Function}
To balance the flexibility of the learnable threshold parameter $\alpha$ and ensure training stability, we introduce a mean squared error-based regularization loss into the optimization objective. This loss constrains deviations of $\alpha$ from its predefined initial value $\alpha_{target}$, preventing parameter drift that could destabilize the node selection process. Specifically, the regularization loss is defined as:
\begin{equation}
L_{\text {reg }}=\lambda \cdot\left(\alpha-\alpha_{\text {target }}\right)^2,
\end{equation}
where $\alpha_{target}$ is initialized to 0.5, responsible for moderating sensitivity to node amplitude distributions, and $\lambda$ is a balancing coefficient determined via cross-validation. This formulation allows $\alpha$ to adapt within a reasonable range while suppressing extreme values. The total loss function combines the cross-entropy and regularization:
\begin{equation}
L_{total}=L_{task} + L_{reg}.
\end{equation}

\section{Experiments}


\subsection{Datasets}
\textbf{NTU-26 Dataset. } NTU-RGBD 120 \cite{21} is a large and widely used 3D human action recognition data collected with Kinect depth sensors, comprising a total of 114,480 samples. Action categories within this dataset are categorized into three main groups: 82 daily actions (such as eating, sitting down, standing up, etc.), 12 health-related actions (including falling down, blowing nose, etc.), and 26 mutual actions (such as hugging, shaking hands, etc). We extract 26 kinds of two-person interactive actions called NTU-26 dataset. Similar to NTU-120, it is structured around two benchmarks: 1) cross-subject (X-Sub) and 2) cross-setup (X-Set). In the cross-subject evaluation, the dataset is split such that half of the actors are used for training and the remaining half for testing. In the cross-setup evaluation, samples with even IDs are assigned to the training set, while samples with odd IDs are used for testing.

\noindent \textbf{SBU-Kinect Interaction Dataset (SBU).} The SBU dataset \cite{22} is a benchmark for two-person human interaction, containing 8 action classes: approaching, departing, kicking, pushing, shaking hands, hugging, exchanging objects, and punching. Each class has around 40 sequences, with ``hugging'' and “shaking hands” having only about 20 sequences each. The dataset includes 282 action videos captured in a lab setting, featuring depth images, color images, and skeleton data. In our experiments, we use the skeleton sequences, where each person is represented by 15 joints with 3D coordinates. For evaluation, we follow the standard protocol outlined in [22], using 5-fold cross-validation on the SBU dataset.

\noindent \textbf{Kinetics-10 Dataset.} The Kinetics dataset \cite{23} contains 300,000 video clips covering 400 categories of human actions, with each category including at least 400 clips. Each clip, approximately ten seconds long, is sourced from a unique YouTube video. The dataset encompasses various action types, including single person action, human-object interactions, and human interactions. 
To evaluate our model, we focus on the human interactions and assess performance using TOP-1 and TOP-5 accuracy. The 10 selected classes include: carrying baby, hugging, kissing, massaging back, massaging feet, massaging legs, massaging person's head, shaking hands, slapping, and tickling.


\begin{table}[h]
\centering
\caption{Comparison of Top-1 accuracy (\%) with state-of-the-art methods on the NTU-26 dataset. }
\vspace{-3mm}
\label{tab1}
\renewcommand{\arraystretch}{1}  
\scalebox{1}{
\begin{tabular}{p{2cm}<{\centering}|p{1.7cm}<{\centering}|p{1.3cm}<{\centering}p{1.3cm}<{\centering}}
\toprule
\multirow{2}{*}{Model}  & \multirow{2}{*}{Year} & \multicolumn{2}{c}{NTU-26} \\
\cline{3-4} 
 & & X-Sub & X-Set\\
\midrule
ST-GCN~\cite{8} & AAAI'18 & 78.90 & 76.10 \\
AS-GCN~\cite{38} & CVPR'19 & 82.90 & 83.70 \\
2s-AGCN~\cite{9} & CVPR'19 & 83.19 & 84.06 \\
MS-G3D~\cite{24} & CVPR'20 & 85.40 & 86.33 \\
DSTA-Net~\cite{44} & CVIPPR'20 & 88.92 &90.10 \\
CTR-GCN~\cite{11} & ICCV'21 & 89.32 & 90.19 \\
LST~\cite{41} & ArXiv'22 & 89.27 & 90.60 \\
TCA-GCN~\cite{42} & ArXiv'22 & 88.37 & 89.30  \\
IGFormer~\cite{5} & ECCV'22  & 85.40 & 86.50 \\
2P-GCN~\cite{3} & TCSVT'22 & 89.56 & - \\
HD-GCN~\cite{43}& ICCV'23 & 88.25 & 90.08 \\
STSA-Net~\cite{37} & Neuroc'23 & 90.20& 90.97 \\
AHNet~\cite{40}  & PR'24  & 86.43  & 86.64 \\
me-GCN~\cite{39} & THMS'25  & 90.00  & 90.00\\
\hline
\textbf{ASEA (Ours)} & - & \textbf{90.52} & \textbf{91.77} \\
\bottomrule
\end{tabular}}
\vspace{-5mm}
\end{table}

\subsection{Comparison with state-of-the-art methods}
We compare our method with state-of-the-art (SOTA) action recognition methods on NTU-26, SBU and Kinetics-10 datasets with only the joint modality. As with previous methods \cite{5, 43}, our experiments follow a fully fair evaluation protocol to ensure unbiased comparison. 

\noindent \textbf{NTU-26.} As shown in Table \ref{tab1}, we compare the Top-1 accuracy of our ASEA model with state-of-the-art methods on the NTU-26 dataset under both X-Sub and X-Set. Our method achieves 90.52\% accuracy on X-Sub and 91.77\% on X-Set, outperforming all previously reported methods. The performance improvement is attributed to the adaptive node selection mechanism, which dynamically identifies the most informative joints, and the external attention module, which captures semantic relationships among active nodes to enhance interaction modeling.

\noindent \textbf{SBU.} We compare the average accuracy of our best results from 5-fold cross-validation with state-of-the-art methods on the SBU dataset, focusing primarily on interaction recognition models. As shown in Table \ref{tab2}, our model achieves accuracy comparable to existing SOTA approaches. This demonstrates the effectiveness of adaptive node selection in learning fine-grained interaction cues, even in datasets with limited samples.

\begin{table}[h]
\centering
\caption{Comparison of average accuracy (\%) over 5-fold cross-validation with state-of-the-art methods on the SBU dataset. }
\vspace{-3mm}
\label{tab2}
\renewcommand{\arraystretch}{1}  
\scalebox{1}{
\begin{tabular}{p{3cm}<{\centering}|p{1.7cm}<{\centering}|p{2cm}<{\centering}}
\toprule
 Model & Year & Accuracy (\%) \\
\midrule
ST-LSTM~\cite{25} & TPAMI'17 & 93.30 \\ 
HCN~\cite{29} & IJCAI'18 & 98.60 \\
VA-fusion~\cite{28} & TPAMI'19 & 98.30 \\ 
K-GCN~\cite{2} & Neuroc'21  &97.20\\
LSTM-IRN~\cite{27} & TMM'21 & 98.20\\
PJD+FCGC~\cite{30} & ICPR'21 & 96.80 \\
GeomNet~\cite{31} & ICCV'21 & 96.33  \\
DR-GCN~\cite{4} & PR'21 & 99.06\\  
AIGCN~\cite{33} & ICME'22 & 99.10\\
2P-GCN~\cite{3} & TCSVT'22 & 98.90\\
IGFormer~\cite{5} & ECCV'22 & 98.40\\
ISTA-Net~\cite{6} & IROS'23  & 98.51\\
AARN~\cite{48}  & CVPR'24  & 98.30\\
\textbf{ASEA (Ours)} &- & \textbf{99.82}\\
\bottomrule
\end{tabular}}
\vspace{-3mm}
\end{table}

\noindent \textbf{Kinetic-10.} Since the Kinetics-10 dataset is a video collection from YouTube, there are some challenges, such as incomplete skeletons on both sides of the interaction. Our model achieves 58.18\% Top-1 and 94.02\% Top-5 accuracy using only the 3D coordinates of joints, outperforming all state-of-the-art methods. Compared to MS-G3D + AEA, the second-best performer, our model improves Top-1 and Top-5 accuracy by 0.60\% and 0.48\%, respectively, while reducing the number of parameters by more than 50\%. Furthermore, by using different baselines in our ASEA method, significant performance improvements are observed across all models. These improvements suggest that our model effectively suppresses noise from missing or occluded joints while preserving essential relational information, demonstrating its robustness in challenging scenarios.

\vspace{-3mm}
\begin{table}[h]
\centering
\caption{Comparison of Top-1 and Top-5 accuracy (\%) after integrating the AEA module into various baseline models on the Kinetics-10 dataset. The AEA module consists of the Adaptive Temporal Node Amplitude Calculation module and the Extra Attention module. }
\vspace{-2mm}
\label{tab3}
\renewcommand{\arraystretch}{1}  
\scalebox{1}{
\begin{tabular}{c|cc|cc}
\toprule
\multirow{2}{*}{Model}  & \multicolumn{2}{c|}{Kinetics-10} & \multirow{2}{*}{\#Param.} & \multirow{2}{*}{FLOPs} \\
\cline{2-3} 
 & Top-1 (\%) & Top-5 (\%) \\
\midrule
ST-GCN~\cite{8} & 49.09  & 87.07 & 3.10M  & 50.14G\\ 
ST-GCN + AEA  & $\textbf{54.34}^{\uparrow\textbf{5.25}}$ & $\textbf{92.03}^{\uparrow\textbf{4.96}}$ & 3.81M  & 61.99G \\
\hline
2s-GCN~\cite{9}  & 54.59 & 92.34 & 3.47M  & 57.30G\\
2s-AGCN + AEA  & $\textbf{56.36}^{\uparrow\textbf{1.77}}$ & $\textbf{{93.33}}^{\uparrow\textbf{{1.30}}}$ & 4.18M  & 69.15G\\ 
\hline
CTR-GCN~\cite{11} & 54.55 & 93.07 & 5.68M  & 33.95G \\
CTR-GCN + AEA & $\textbf{56.57}^{\uparrow\textbf{2.02}}$ & $\textbf{93.41}^{\uparrow\textbf{0.34}}$ & 6.42M &45.70G\\
\hline
MS-G3D~\cite{24} & 56.16 & 93.13 &3.14M & 79.26G \\
MS-G3D + AEA  & $\textbf{57.58}^{\uparrow\textbf{1.42}}$ & $\textbf{93.54}^{\uparrow\textbf{0.41}}$ & 4.66M & 105.89G\\  
\hline
DeGCN \cite{47}  & 50.10 & 90.51 & 1.32M & 174.79G\\
DeGCN + AEA & $\textbf{51.51}^{\uparrow\textbf{1.41}}$ & $\textbf{90.72}^{\uparrow\textbf{2.10}}$ & 2.06M  & 263.69G \\
\hline
\textbf{ASEA (Ours)} & \textbf{58.18} & \textbf{94.02}  & \textbf{2.04M}  &\textbf{30.82G}\\
\bottomrule
\end{tabular}}
\vspace{-3mm}
\end{table}



\begin{table}[h]
\centering
\caption{Performance comparison with different modules. }
\vspace{-3mm}
\label{tab4}
\renewcommand{\arraystretch}{1}  
\scalebox{1}{
\begin{tabular}{c|cc}
\toprule
\multirow{2}{*}{Model} & \multicolumn{2}{c}{NTU-26}\\
\cline{2-3} 
 & X-Sub (\%) & X-Set (\%) \\
\midrule
Baseline &89.55 &90.83 \\
ASEA w/o AT-NAC & 89.79 & 91.50 \\ 
ASEA & \textbf{90.52} & \textbf{91.77} \\
\bottomrule
\end{tabular}}
\vspace{-3mm}
\end{table}

\begin{figure}[htbp]
    \centering
    \subfloat[]{\includegraphics[width=0.23\linewidth]{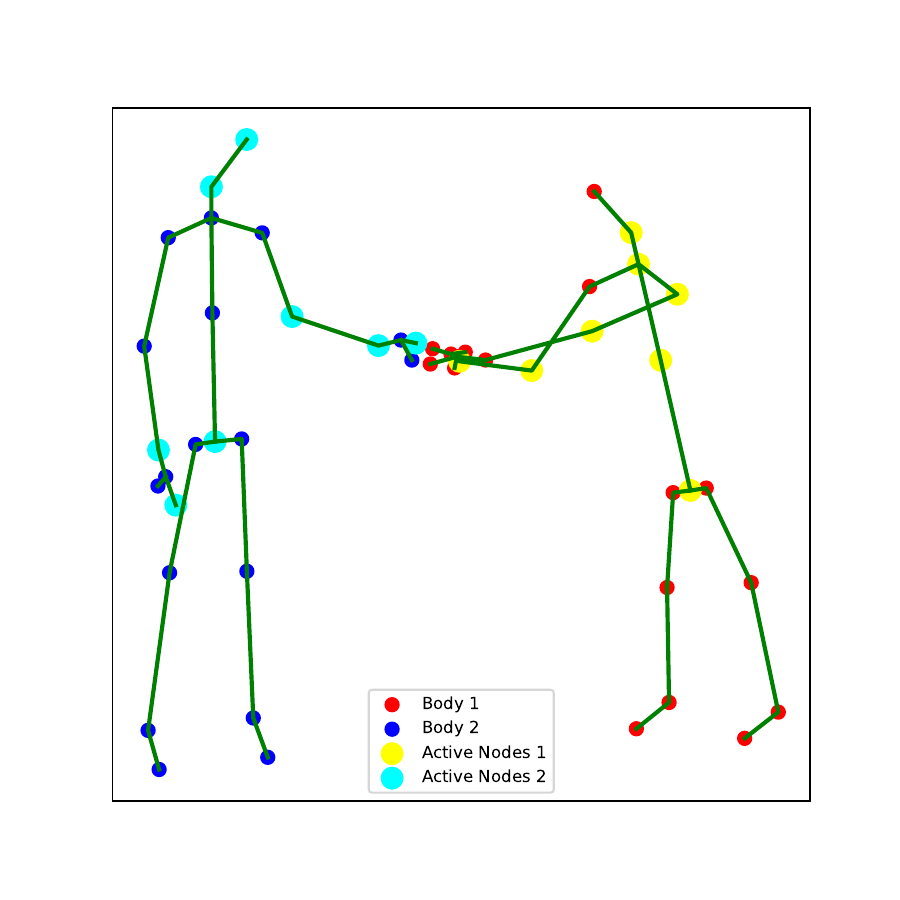}\label{fig:sub1}}
    \hspace{1mm}
    \subfloat[]{\includegraphics[width=0.23\linewidth]{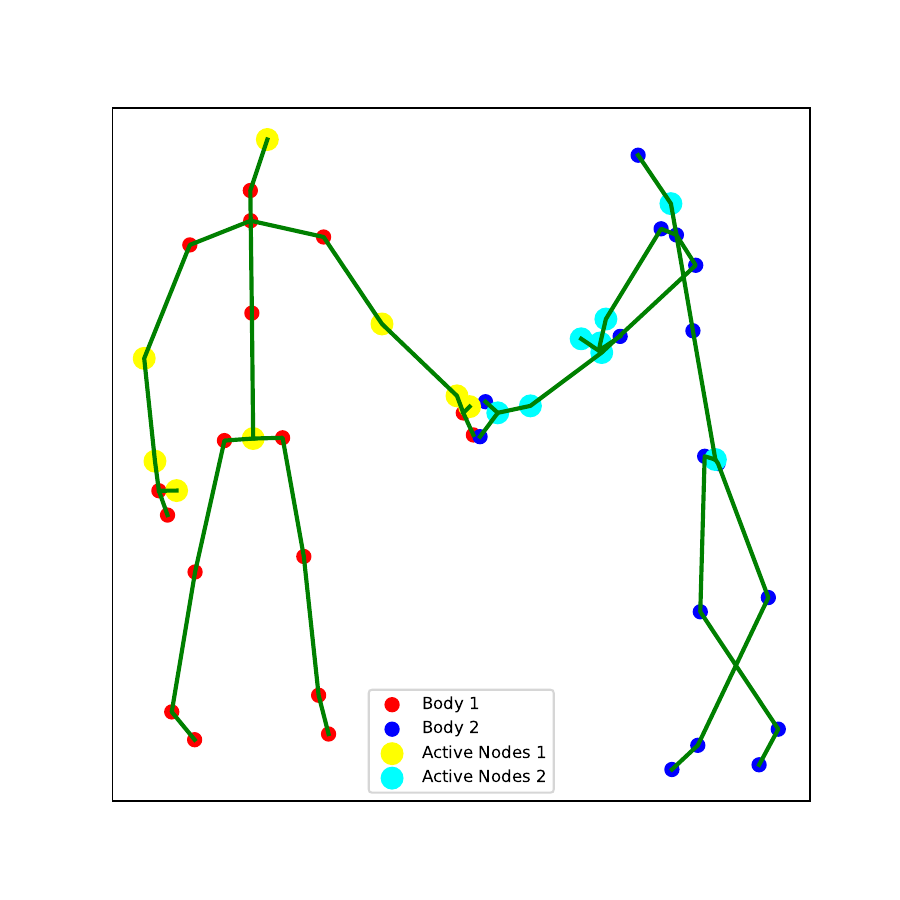}\label{fig:sub2}}
    \hspace{1mm}
    \subfloat[]{\includegraphics[width=0.228\linewidth]{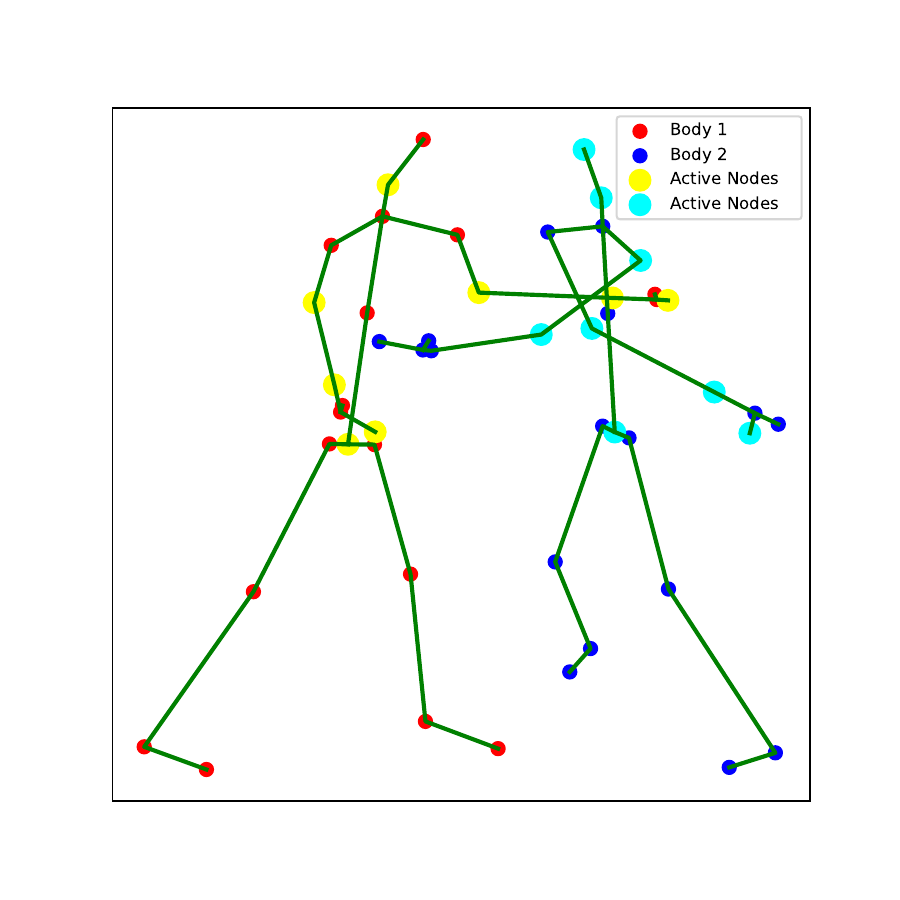}\label{fig:sub3}}
    \hspace{1mm}
    \subfloat[]{\includegraphics[width=0.225\linewidth]{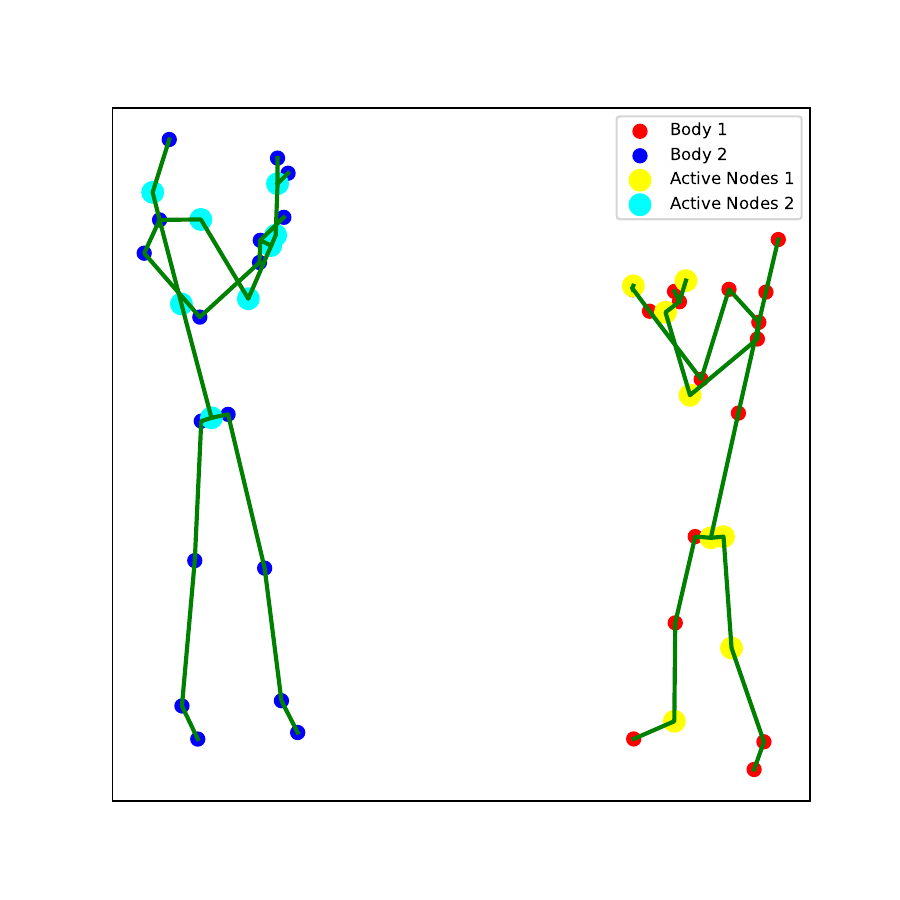}\label{fig:sub4}}
    \vspace{-3mm}
    \caption{Visualization of the selection of active nodes across different actions. (a) Cheers, (b) Shaking hands, (c) Hugging, and (d) Taking photos. }
    \label{fig3}
    \vspace{-5mm}
\end{figure}

\subsection{Ablation experiments}
We conduct detailed ablation experiments on the NTU-26 dataset using the joint modality. 

\begin{figure}[]
    \centering
    \begin{subfigure}[b]{0.456\linewidth}
        \centering
        \includegraphics[width=\linewidth]{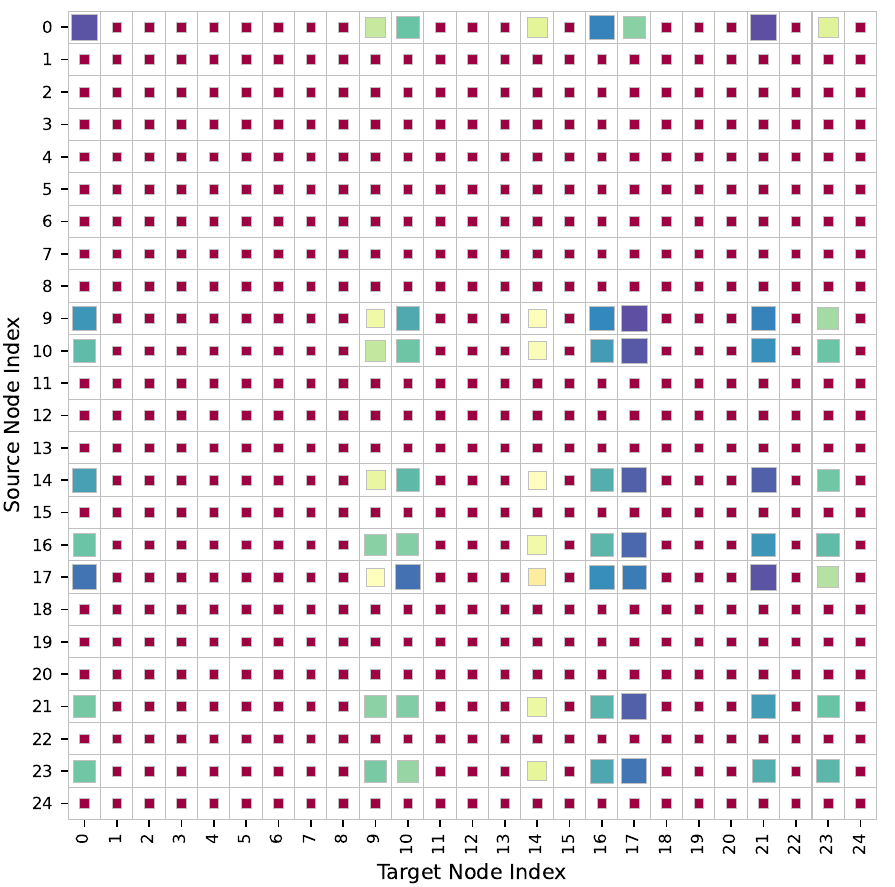}
        \caption{}
        \label{fig:sub1}
    \end{subfigure}
    \hfill
    \begin{subfigure}[b]{0.53\linewidth}
        \centering
        \includegraphics[width=\linewidth]{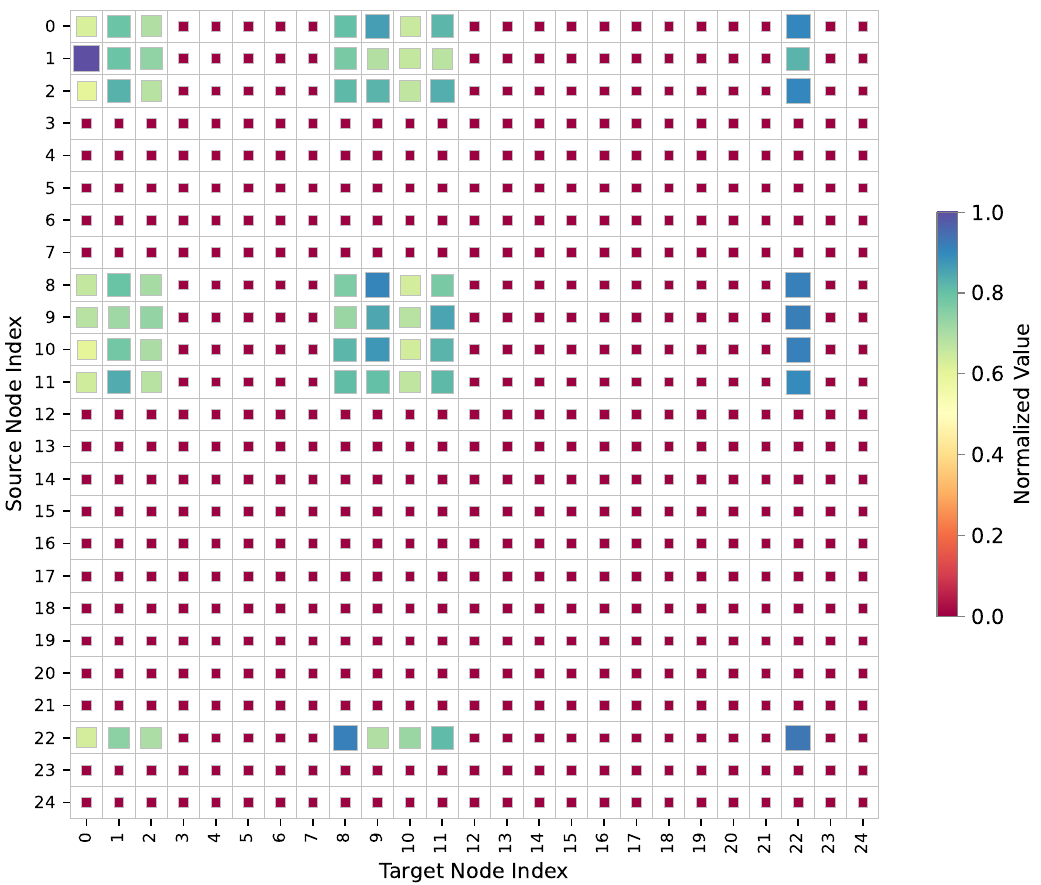}
        \caption{}
        \label{fig:sub2}
    \end{subfigure}
    \vspace{-7mm}
    \caption{Visualization of attention patterns for the ``taking photos'' action. (a) Attention visualization with body 1 as the first-person perspective. (b) Attention visualization with body 2 as the first-person perspective. }
    \label{fig4}
    \vspace{-3
    mm}
\end{figure}

Table \ref{tab4} presents a comparative analysis of employing different modules of our method. We removed the Adaptive Temporal Node Amplitude Calculation (AT-NC) module and applied external attention across all nodes of different individuals. This led to an accuracy improvement over the baseline, demonstrating the effectiveness of external attention in capturing inter-individual interactions by modeling relationships between different individuals. Nevertheless, the absence of the proposed AT-NAC module led to a decline in accuracy compared to ASEA, underscoring the critical role of AT-NAC in selectively identifying the most relevant and active nodes. This selective mechanism not only enhances the model's ability to capture meaningful interactions but also mitigates noise, thereby improving the overall model efficiency and performance. To provide deeper insights, we visualized the selection of active nodes across different actions, as shown in Fig. \ref{fig3}. 
The visualization in Fig. \ref{fig3} demonstrates that while actions such as handshaking, cheering, and hugging exhibit similar active node patterns, each action shows subtle differences. Handshaking primarily involves the wrists and elbows, while cheering and hugging emphasize the shoulders, with some overlap in the torso region. In contrast, the action of taking a photo introduces a shift in node selection, extending to the legs for stabilizing and adjusting the camera. This variation in selected nodes across different actions highlights the dynamic and context-specific nature of the proposed AT-NAC module, enabling the model to adaptively capture key interaction dynamics based on the action context.
Furthermore, to illustrate the capability of external attention in capturing implicit semantic information beyond geometric relationships, we visualized the attention patterns using the ``taking photos'' action as a representative scenario, where interactions primarily rely on gaze direction and gestures rather than physical contact, as shown in Fig. \ref{fig4}.

\begin{table}[h]
\centering
\caption{Comparison of the results obtained by selecting active nodes based on node velocity and the temporal weighted $L_2$-norm of learned features on the NTU-26 dataset. }
\vspace{-3mm}
\label{tab5}
\renewcommand{\arraystretch}{1}  
\scalebox{1}{
\begin{tabular}{c|cc}
\toprule
\multirow{2}{*}{Model} & \multicolumn{2}{c}{NTU-26}  \\
\cline{2-3} 
 & X-Sub (\%) & X-Set (\%)\\
\midrule
ASEA w/node velocity & 76.35 & 82.76 \\ 
ASEA & \textbf{90.52} & \textbf{91.77}\\
\bottomrule
\end{tabular}}
\vspace{-3mm}
\end{table}

\noindent \textbf{Active Node Selection Strategy.} To verify the effectiveness of selecting active nodes based on the temporal weighted $L_2$-norm of learned features, we conducted a comparative study with the approach of selecting active nodes based on node velocity. This experiment focused on evaluating the consistency of capturing key nodes and maintaining intra-class consistency across different individuals performing the same action. As shown in Table \ref{tab5}, ASEA w/node velocity experiences a significant drop in accuracy compared to our method. 

\begin{figure}[htbp]
    \centering
    \begin{subfigure}[b]{0.435\linewidth}
        \centering
        \includegraphics[width=\linewidth]{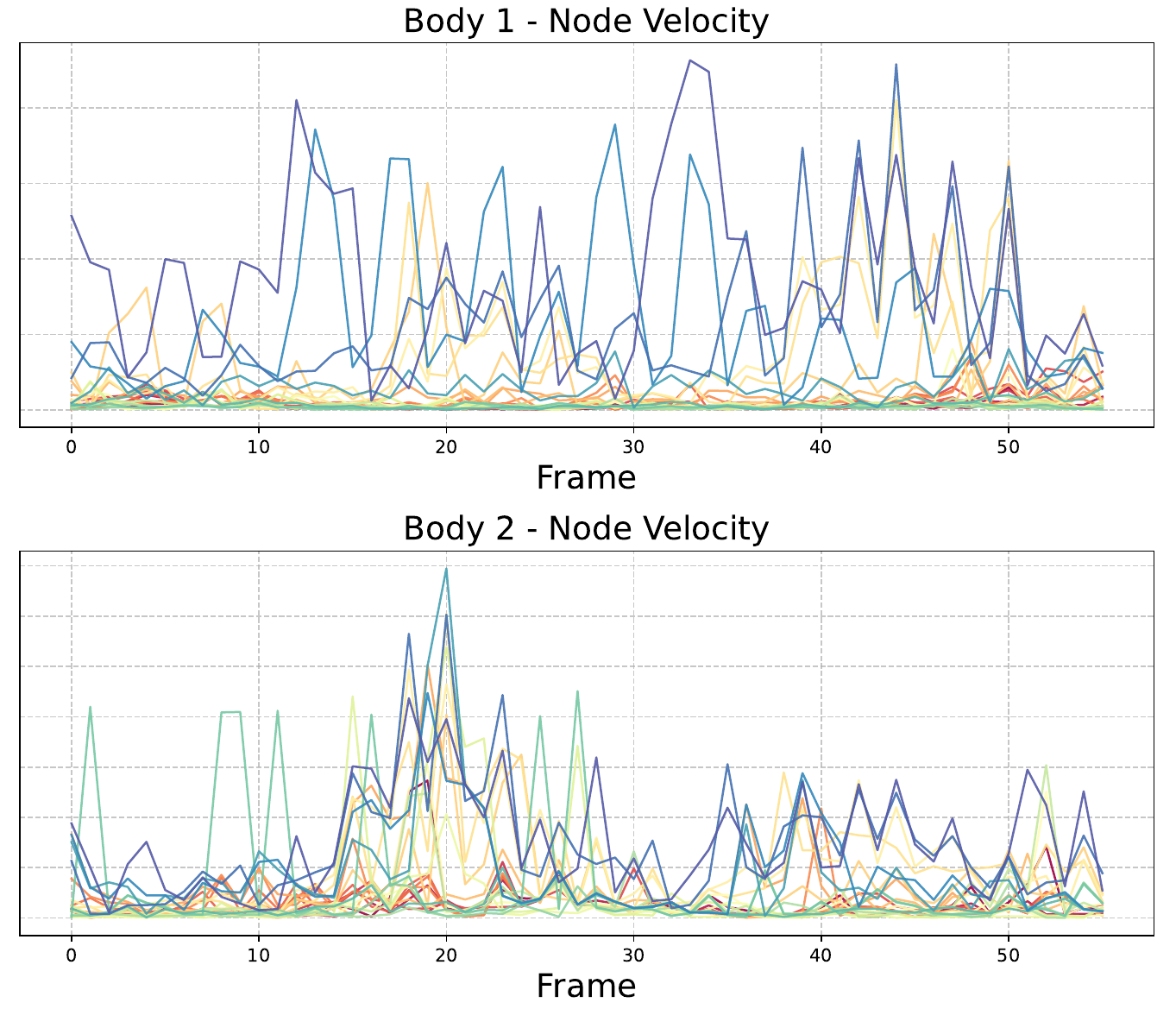}
        \vspace{-5mm}
        \caption{}
        \label{fig:sub1}
    \end{subfigure}
    \hfill
    \begin{subfigure}[b]{0.543\linewidth}
        \centering
        \includegraphics[width=\linewidth]{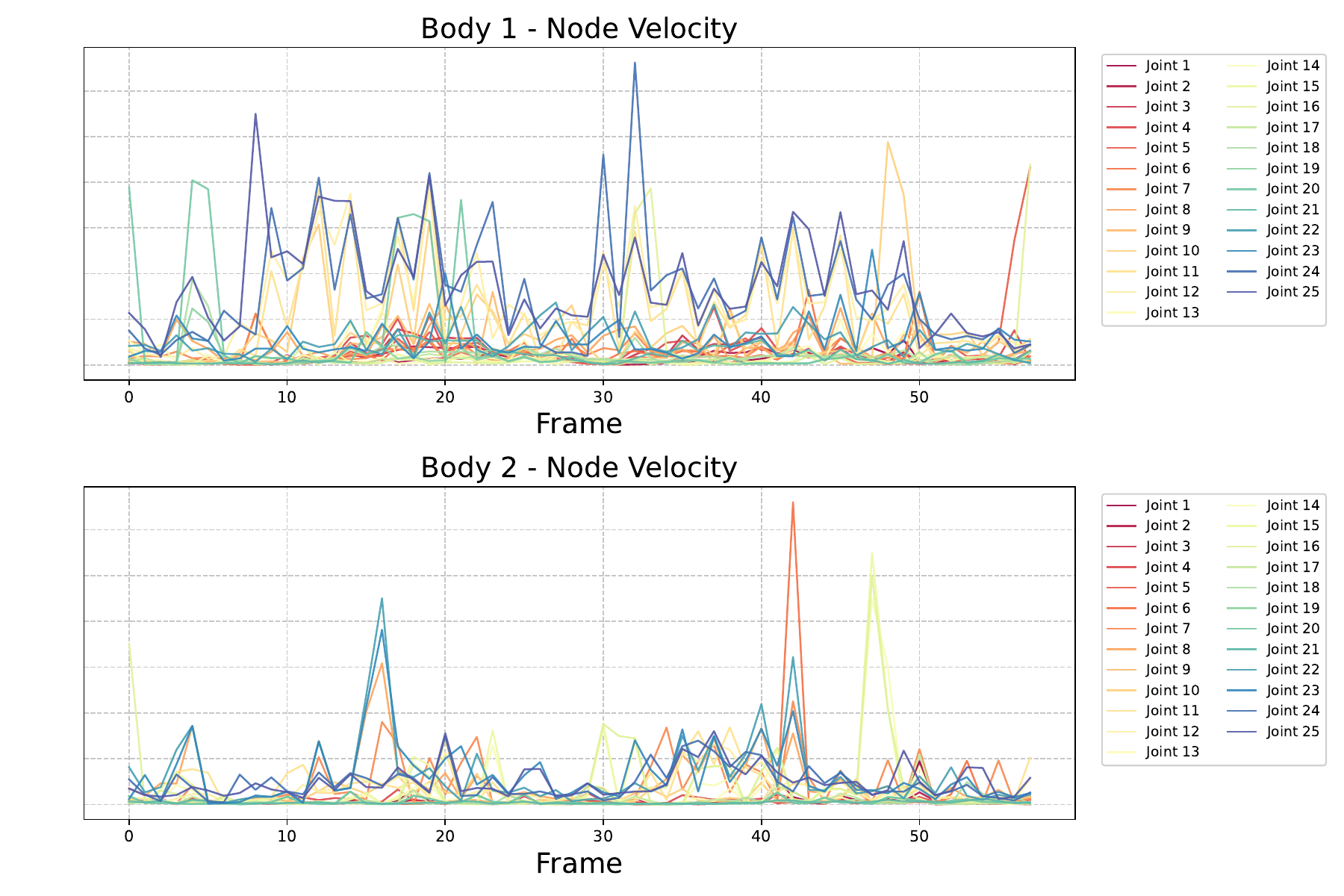}
        \vspace{-5mm}
        \caption{}
        \label{fig:sub2}
    \end{subfigure}
    \\
    \begin{subfigure}[b]{0.435\linewidth}
        \centering
        \includegraphics[width=\linewidth]{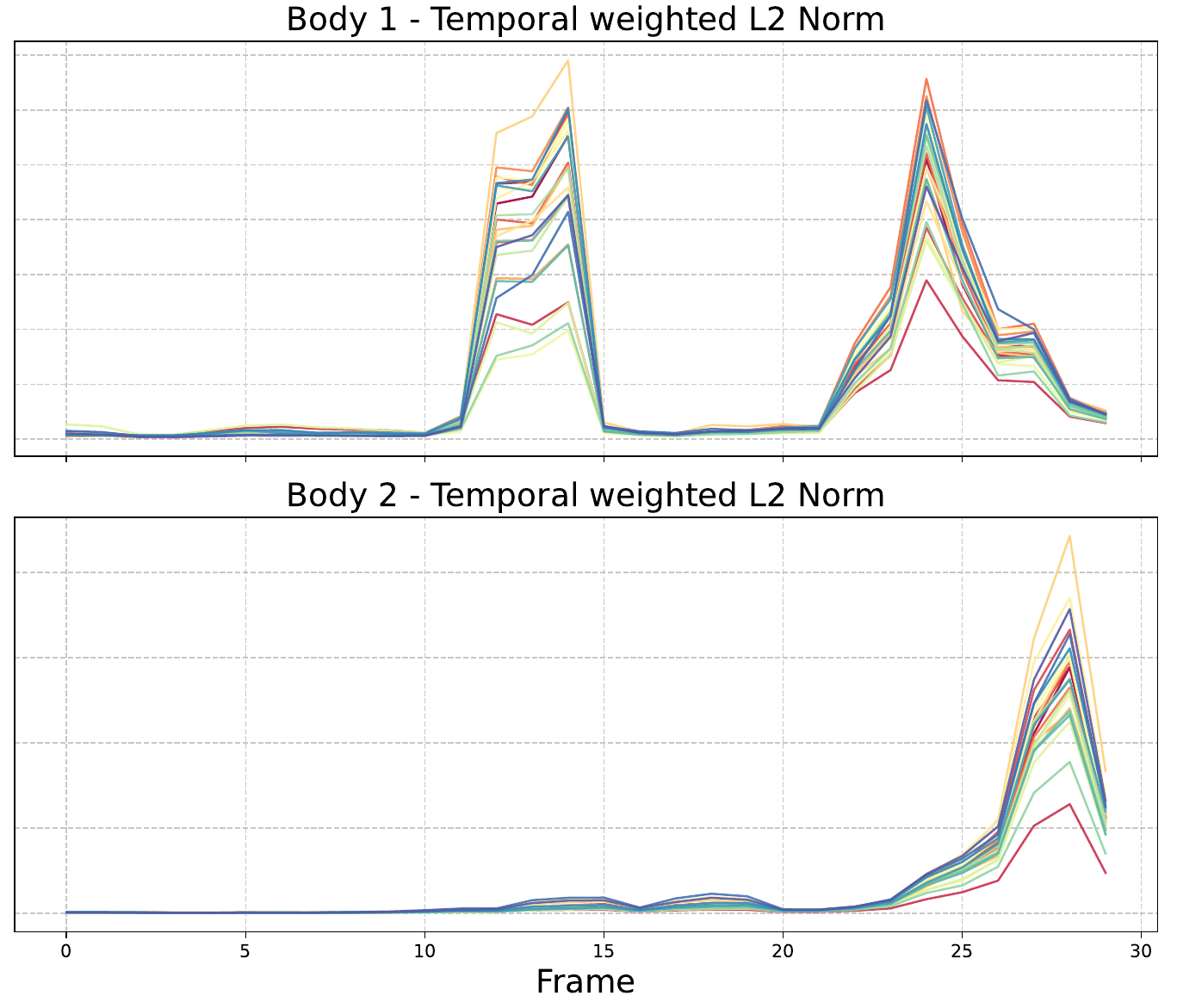}
        \caption{}
        \label{fig:sub1}
    \end{subfigure}
    \hfill
    \begin{subfigure}[b]{0.535\linewidth}
        \centering
        \includegraphics[width=\linewidth]{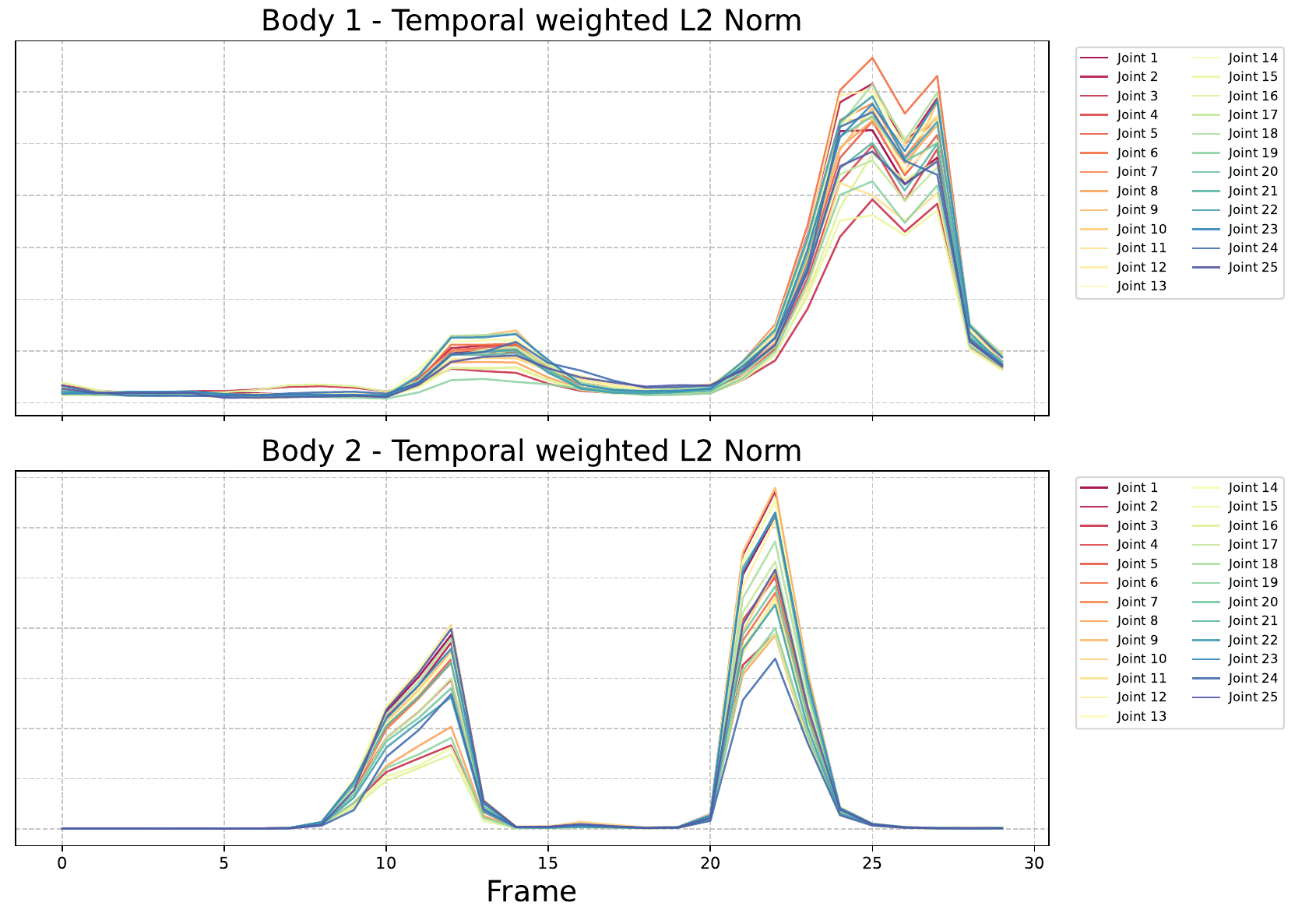}
        \caption{}
        \label{fig:sub2}
    \end{subfigure}
    \vspace{-3mm}
    \caption{Visualization of  node variations for different samples of the same action category. (a) and (b) show the velocity curves of all nodes for different samples. (c) and (d) illustrate the corresponding temporal weighted $L_2$-norm curves. }
    \label{fig5}
    \vspace{-5mm}
\end{figure}

To further analyze the reasons for this performance degradation, we visualized the node variations for different samples of the same action category in Fig. \ref{fig5} (e.g., cheers), comparing the results of using velocity and the temporal weighted $L_2$-norm of learned features. Fig. \ref{fig5} (a) and (b) show the velocity curves of all nodes for different samples, while (c) and (d) illustrate the corresponding temporal weighted $L_2$-norm curves. By comparing the node curves intersected by the vertical lines at the peak regions,
we observe that the velocity curves exhibit significant variations across different samples, whereas the temporal weighted $L_2$-norm curves remain highly consistent. The visualizations reveal that selecting active nodes based on node velocity increases intra-class variations and reduces recognition accuracy due to its inability to capture global motion patterns. Node velocity primarily reflects short-term frame-to-frame changes, which introduces inconsistencies across individuals performing the same action, leading to greater intra-class variations. In contrast, our approach, which selects active nodes based on the temporal weighted $L_2$-norm of the learned features, significantly reduces intra-class differences and improves the overall consistency of the model's performance across different individuals.

\section{Conclusion}
In this paper, we proposed a novel approach, the Active Node Selection with External Attention Network (ASEA), to dynamically capture interaction relationships without relying on predefined assumptions. Specifically, we introduce Graph Convolution Network (GCN) to model the intra-personal relationships of each participant, enabling a fine-grained representation of individual actions. To identify the most informative nodes, we design the Adaptive Temporal Node Amplitude Calculation (AT-NAC) module, which dynamically assigns temporal weights to each frame based on its significance and computes node amplitude to reflect spatial and temporal variations. During this process, a learnable threshold, regularized to maintain a balanced range of node amplitudes, prevents extreme deviations. To further capture interactions between individuals, we apply the External Attention (EA) module to the active nodes across different individuals, allowing the model to focus on interaction-critical regions while suppressing irrelevant information. Comprehensive experiments conducted on multiple public datasets demonstrate that our method achieves superior performance compared to state-of-the-art methods. Ablation studies further validate its effectiveness in capturing dynamic interaction patterns and filtering out irrelevant information, highlighting its superiority in interaction recognition. 

\bibliographystyle{ACM-Reference-Format}
\bibliography{sample-base}

\appendix









\end{document}